# A Logic for Reasoning about Evidence

**Joseph Y. Halpern**
Department of Computer Science
Cornell University
Ithaca, NY 14853
halpern@cs.cornell.edu
http://www.cs.cornell.edu/home/halpern

**Riccardo Pucella**
Department of Computer Science
Cornell University
Ithaca, NY 14853
riccardo@cs.cornell.edu

## Abstract

We introduce a logic for reasoning about evidence, that essentially views evidence as a function from prior beliefs (before making an observation) to posterior beliefs (after making the observation). We provide a sound and complete axiomatization for the logic, and consider the complexity of the decision problem. Although the reasoning in the logic is mainly propositional, we allow variables representing numbers and quantification over them. This expressive power seems necessary to capture important properties of evidence.

## 1 Introduction

Consider the following situation, essentially taken from [Halpern and Tuttle 1993; Fagin and Halpern 1994]. A coin is tossed, which is either fair or double-headed. The coin lands heads. How likely is it that the coin is double-headed? What if the coin is tossed 20 times and it lands heads each time? Intuitively, it is much more likely that the coin is double-headed in the latter case than in the former. But how should the likelihood be measured? A straightforward application of probability theory, perhaps the best candidate, is not possible here. We cannot compute, for instance, the probability of the coin being double-headed; assigning a probability to that event requires that we have a prior probability on the coin being double-headed. For example, if the coin was chosen at random from a barrel with one billion fair coins and one double-headed coin, it is still overwhelmingly likely that the coin is fair, and that the sequence of 20 heads is just unlucky. However, in the problem statement, there is no prior probability mentioned. We could of course posit a prior probability and see how the posterior probability behaves when we change the prior, but the point is that our intuition does not seem to rely on such a posited prior.

The main feature of this situation is that it involves a combination of probabilistic outcomes (e.g., the coin tosses) and nonprobabilistic outcomes (e.g., the choice of the coin). There has been a lot of work on reasoning about systems that combine both probabilistic and nondeterministic choices (e.g., [Vardi 1985; Fischer and Zuck 1988; Halpern, Moses, and Tuttle 1988; van Glabbeek, Smolka, Steffen, and Tofts 1990; Larsen and Skou 1991; Halpern and Tuttle 1993; de Alfaro 1998]). However, the observations above suggest that if we attempt to formally analyze this situation in one of those frameworks, which essentially permit only the modeling of probabilities, we will not be able to directly capture this intuition about increasing likelihood. To see how this plays out, consider a formal analysis of the situation in the Halpern-Tuttle [1993] framework. Suppose that Alice nonprobabilistically chooses one of two coins: a fair coin with probability 1/2 of landing heads, or a double-headed coin with probability 1 of landing heads. Alice tosses this coin repeatedly. Let $\varphi_k$ be a formula stating: "the $k$th coin toss lands heads". What is the probability of $\varphi_k$ according to Bob, who does not know which coin Alice chose?

According to the Halpern-Tuttle framework, this can be modeled by considering the set of runs describing the states of the system at each point in time, and partitioning this set into two subsets, one for each coin used. In the set of runs where the fair coin is used, the probability of $\varphi_k$ is $1/2$; in the set of runs where the double-headed coin is used, the probability of $\varphi_k$ is 1. In this setting, the only conclusion that can be drawn is $(\Pr_B(\varphi_k) = 1/2) \vee (\Pr_B(\varphi_k) = 1)$. (This is of course the probability from Bob's point of view; Alice presumably knows which coin she is using.) Intuitively, this seems reasonable: if the fair coin is chosen, the probability that the $k$th coin toss lands heads, according to Bob, is 1/2; if the double-headed coin is chosen, the probability is 1. Since Bob does not know which of the coins is being used, that is all that can be said.

But now suppose that, before the 101st coin toss, Bob learns the result of the first 100 tosses. Suppose, moreover, that all of these landed heads. What is the probability that the 101st coin toss lands heads? By the same analysis, it is still either 1/2 or 1, depending on which coin is used.

This is hardly useful. To make matters worse, no matter how many coin tosses Bob witnesses, the probability that the next toss lands heads remains unchanged. But this answer misses out on some important information. The fact that all of the first 100 coin tosses are heads is very strong *evidence* that the coin is in fact double-headed. Indeed, a straightforward computation using Bayes' Rule shows that if the prior probability of the coin being double-headed is $\alpha$, then after observing that all of the 100 tosses land heads, the probability of the coin being double-headed becomes

$$\frac{\alpha}{\alpha + 2^{-100}(1-\alpha)} = \frac{2^{100}\alpha}{2^{100}\alpha + (1-\alpha)}.$$

However, note that it is not possible to determine the posterior probability that the coin is double-headed (or that the 101st coin toss is heads) without the prior probability $\alpha$. After all, if Alice chooses the double-headed coin with probability only $10^{-100}$, then it is still overwhelmingly likely that the coin used is in fact fair, and that Bob was just very unlucky to see such a nonrepresentative sequence of coin tosses.

We are not aware of any framework for reasoning about nondeterminism and probability that takes the issue of evidence into account. On the other hand, evidence has been discussed extensively in the philosophical literature. Much of this discussion occurs in the philosophy of science, specifically *confirmation theory*, where the concern has been historically to assess the support that evidence obtained through experimentation lends to various scientific theories [Carnap 1962; Popper 1959; Good 1950; Milne 1996]. (Kyburg [1983] provides a good overview of the literature.)

In this paper, we introduce a logic for reasoning about evidence. Our logic extends a logic defined by Fagin, Halpern and Meggido [1990] (FHM from now on) for reasoning about likelihood expressed as either probability or belief. The logic has first-order quantification over the reals (so includes the theory of real closed fields), as does the FHM logic, for reasons that will shortly become clear. We add observations to the states, and refine the language to talk about both the prior probability of hypotheses and the posterior probability of hypotheses, taking into consideration the observation at the states. We provide an additional operator to talk about the evidence carried by particular observations. This lets us write formulas that talk about the relationship between the prior and posterior probabilities, and the evidence of the observations.

We then provide a sound and complete axiomatization for the logic. To obtain such an axiomatization, we seem to need first-order quantification in a fundamental way. Roughly speaking, this is because ensuring that the evidence operator has the appropriate properties requires us to assert the existence of suitable probability measures. It does not seem possible to do this without existential quantification. Finally, we consider the complexity of the satisfiability problem. The satisfiability problem for the full language requires doubly-exponential time, since it incorporates the theory of real closed fields, for which such a lower bound is known [Weispfenning 1988]. However, we show that the satisfiability problem for a propositional fragment of the language, which is still strong enough to allow us to express many properties of interest, is NP-complete.

The rest of the paper is organized as follows. In the next section, we formalize a notion of evidence that captures the intuitions outlined above. In Section 3, we introduce our logic for reasoning about evidence. In Section 4, we present an axiomatization for the logic and show that it is sound and complete with respect to the intended models. In Section 5, we discuss the complexity of the decision problem of our logic.

## 2  Measures of confirmation and evidence

In order to develop a logic for reasoning about evidence, we need to first formalize an appropriate notion of evidence. Evidence has been studied in depth in the philosophical literature, under the name of *confirmation theory*. Confirmation theory aims at determining and measuring the support a piece of evidence provides a hypothesis. As we mentioned in the introduction, many different measures of confirmation have been proposed in the literature. Typically, a proposal has been judged on the degree to which it satisfies various properties that are considered appropriate for confirmation. For example, it may be required that a piece of evidence $e$ confirms a hypothesis $h$ if and only if $e$ makes $h$ more probable. We have no desire to enter the debate as to which class of measures of confirmation is more appropriate. For our purposes, most confirmation functions are useless: they assume that we have a prior on the hypotheses, which is exactly the information we do not have and do not want to assume. One exception is measures of evidence that use the log-likelihood ratio. Given an observation $ob$, the degree of confirmation that it provides for a hypothesis $h$ is

$$l(ob, h) = \log\left(\frac{\Pr(ob \mid h)}{\Pr(ob \mid \overline{h})}\right).$$

That is, the degree of confirmation is the ratio is between the probability of observing $ob$ when $h$ holds versus the probability of observing $ob$ when $h$ does not hold. (The notation $\overline{h}$ represents the case where hypothesis $h$ does not hold.) A logarithm is used to ensure that the measure is positive when it confirms the hypothesis. This approach has been advocated by Good [1950, 1960], among others.[1]

---

[1] In the literature, confirmation is usually taken with respect to some background knowledge. For ease of exposition, we ignore background knowledge here, although it can easily be incorpo-

One problem with the log-likelihood ratio measure $l$, as we have defined it, is that it can only be used to reason about evidence discriminating between two competing hypotheses, namely between an hypothesis $h$ holding and the hypothesis $h$ not holding. We would like a measure of confirmation along the lines of the log-likelihood ratio measure, but that can handle multiple competing hypotheses. One such generalization was given by Shafer [1982], in the context of the Dempster-Shafer theory of evidence based on belief functions [Shafer 1976], and studied by Walley [1987]. The description here is taken mostly from [Halpern and Fagin 1992].

We start with a set $\mathcal{H} = \{h_1, \ldots, h_n\}$ of mutually exclusive and exhaustive hypotheses; thus, exactly one hypothesis holds at any given time. Assume a finite set $\mathcal{O}$ of observations (or pieces of evidence), and assume that for each hypotheses $h \in \mathcal{H}$ we have a probability space $(\mathcal{O}, 2^\mathcal{O}, \mu_h)$ that assigns to every observation $ob \in \mathcal{O}$ the probability of $ob$ given that hypothesis $h$ holds. Furthermore, we assume that the observations in $\mathcal{O}$ are relevant to the hypotheses: for every observation $ob \in \mathcal{O}$, there must be an hypothesis $h$ such that $\mu_h(ob) > 0$. We define an evidence space (over $\mathcal{H}$ and $\mathcal{O}$) to be a tuple $\mathcal{E} = (\mathcal{H}, \mathcal{O}, \mu_{h_1}, \ldots, \mu_{h_n})$.

Given an evidence space $\mathcal{E}$, we define the weight that the observation $ob$ lends to hypothesis $h$, written $w_\mathcal{E}(ob, h)$ or simply $w(ob, h)$ when $\mathcal{E}$ is understood, as

$$w_\mathcal{E}(ob, h) = \frac{\mu_h(ob)}{\sum_{h' \in \mathcal{H}} \mu_{h'}(ob)} \quad (1)$$

For a set of hypotheses $H$, we define $w_\mathcal{E}(ob, H)$ as simply the sum of the individual $w_\mathcal{E}(ob, h)$ for $h \in H$. This definition makes $w_\mathcal{E}(ob, \cdot)$ a probability measure on hypotheses, for each fixed observation $ob$. Intuitively, if $w_\mathcal{E}(ob, h) = 1$, then $ob$ fully confirms $h$ (i.e., $h$ is certainly true if $ob$ is observed), while if $w_\mathcal{E}(ob, h) = 0$, then $ob$ disconfirms $h$ (i.e., $h$ is certainly false if $ob$ is observed). We can verify that this is a generalization of the log-likelihood ratio measure. More precisely, given an evidence space $\mathcal{E}$ with $|\mathcal{H}| = 2$, then for a fixed observation $ob$, $w_\mathcal{E}(ob, \cdot)$ induces the same relative order on hypotheses as $l(ob, \cdot)$, and for a fixed hypothesis $h$, $w_\mathcal{E}(\cdot, h)$ induces the same relative order on observations as $l(\cdot, h)$. That is, for all $ob$, $ob'$, $h$, and $h'$, $w_\mathcal{E}(ob, h) \geq w_\mathcal{E}(ob, h')$ if and only if $l(ob, h) \geq l(ob, h')$ and $w_\mathcal{E}(ob, h) \geq w_\mathcal{E}(ob', h)$ if and only if $l(ob, h) \geq l(ob', h)$. Intuitively, the weight $w_\mathcal{E}(ob, h)$ is the probability that $h$ is the right hypothesis in the light of observation $ob$.[2] The advantages of $w_\mathcal{E}$ over other known measures of confirmation are (a) that it it applicable when there is no prior probability distribution on the hypotheses, (b) that it is applicable when there are more than two competing hypotheses, and (c) that it has a fairly intuitive probabilistic interpretation.[3]

Although $w_\mathcal{E}(ob, \cdot)$ behaves like a probability measure on hypotheses for every observation $ob$, it is perhaps best not to think of it as a probability. Rather, it is an encoding of evidence. But what is evidence? Halpern and Fagin [1992] have suggested that evidence can be thought of as a *function* mapping a prior probability on the hypotheses to a posterior probability, based on the piece of evidence witnessed. The weight $w_\mathcal{E}$ can be thought of as representing such a function. In fact, as pointed out by Halpern and Fagin [1992], $w_\mathcal{E}(ob, \cdot)$ can be used to "update" a prior probability $\mu_0$ on the hypotheses $\mathcal{H}$ into a probability $\mu_{ob}$, after observing $ob$ by applying Dempster's Rule of Combination [Shafer 1976]. That is,

$$\mu_{ob} = \mu_0 \oplus w_\mathcal{E}(ob, \cdot), \quad (2)$$

where $\oplus$ combines two probability distributions on $\mathcal{H}$ to get a new probability distribution on $\mathcal{H}$ defined as follows:

$$(\mu_1 \oplus \mu_2)(H) = \frac{\sum_{h \in H} \mu_1(h)\mu_2(h)}{\sum_{h \in \mathcal{H}} \mu_1(h)\mu_2(h)}.$$

Essentially, Dempster's Rule is simulating the effects of Bayes' Rule here.

**Example 2.1:** To get a feel for how this measure of evidence can be used, consider a variation of the two-coins example in the introduction. Assume that the coin chosen by Alice is either double-headed or fair, and consider sequences of hundred tosses of that coin. Let $\mathcal{O} = \{m : 0 \leq m \leq 100\}$ (the number of heads observed), and let $\mathcal{H} = \{F, D\}$, where $F$ is "the coin is fair", and $D$ is "the coin is double-headed". The probability spaces associated with the hypotheses are generated by the following probabilities for simple observations $m$:

$$\mu_F(m) = \frac{1}{2^{100}}\binom{100}{m} \qquad \mu_D(m) = \begin{cases} 1 & \text{if } m = 100 \\ 0 & \text{otherwise.} \end{cases}$$

(We extend by additivity to the whole space $\mathcal{O}$.) For any

---

rated into the framework we present.

[2]We could have taken the log of the ratio to make $w_\mathcal{E}$ more in line with the log-likelihood ratio $l$ defined earlier, but there are technical advantages in having the weight of evidence be a number between 0 and 1.

[3]Another representation that has similar characteristics is Shafer's original representation of evidence via belief functions [Shafer 1976], given essentially by

$$w(ob, H) = \frac{\max_{h \in H} \mu_h(ob)}{\max_{h \in \mathcal{H}} \mu_h(ob)}.$$

This measure is known in statistical hypothesis testing as the *generalized likelihood-ratio statistic*. It is another generalization of the log-likelihood ratio measure $l$. At this point, one may well ask what could help decide which weight function to use. In the case of these two particular notions of evidence, their main difference is in how they behave when one considers the combination of evidence. Arguably, the measure that we have used behaves better in this case [Walley 1987; Halpern and Fagin 1992].

observation $m \neq 100$, the weight in favor of $F$ is given by

$$w(m, F) = \frac{\frac{1}{2^{100}}\binom{100}{m}}{0 + \frac{1}{2^{100}}\binom{100}{m}} = 1,$$

which intuitively means that the support of $m$ is unconditionally provided to $F$; indeed, any such sequence of tosses cannot appear with the double-headed coin. Thus, if $m > 0$, we get that

$$w(m, D) = \frac{0}{0 + \frac{1}{2^{100}}\binom{100}{m}} = 0.$$

What happens when the hundred coin tosses are all heads? It is straightforward to check that

$$w(100, F) = \frac{\frac{1}{2^{100}}}{1 + \frac{1}{2^{100}}} = \frac{1}{1 + 2^{100}}$$

$$w(100, D) = \frac{1}{1 + \frac{1}{2^{100}}} = \frac{2^{100}}{1 + 2^{100}};$$

this time there is overwhelmingly more evidence in favor of $D$ than $F$.

Note that we have not assumed any prior probability. Thus, we cannot talk about the probability that the coin is fair or double-headed. What we have is a quantitative assessment of the evidence in favor of one of the hypotheses. However, if we assume a prior probability $\alpha$ on the coin being fair and $m$ heads are observed after 100 tosses, then the probability that the coin is fair is 1 if $m \neq 100$; if $m = 100$ then, applying the rule of combination, the posterior probability of the coin being fair is $\alpha/(\alpha + (1-\alpha)2^{100})$. ∎

Can we characterize weight functions using a small number of properties? More precisely, given sets $\mathcal{H}$ and $\mathcal{O}$, and a function $f$ from $\mathcal{O} \times \mathcal{H}$ to $[0,1]$, can we give properties of $f$ that ensure that there are probability measures $\{\mu_h\}_{h \in \mathcal{H}}$ such that $f = w_\mathcal{E}$? As we saw before, for a fixed observation $ob$, $f$ acts like a probability measure on $\mathcal{H}$. However, this is not sufficient to guarantee that $f$ is a weight function. Consider the following example, with $\mathcal{O} = \{ob_1, ob_2\}$ and $\mathcal{H} = \{h_1, h_2, h_3\}$:

$f(ob_1, h_1) = 1/4 \qquad f(ob_2, h_1) = 1/4$
$f(ob_1, h_2) = 1/4 \qquad f(ob_2, h_2) = 1/2$
$f(ob_1, h_3) = 1/2 \qquad f(ob_2, h_3) = 1/4.$

It is straightforward to check that $f(ob_1, \cdot)$ and $f(ob_2, \cdot)$ are probability measures on $\mathcal{H}$, but that there is no evidence space $\mathcal{E} = (\mathcal{H}, \mathcal{O}, \mu_{h_1}, \mu_{h_2}, \mu_{h_3})$ such that $f = w_\mathcal{E}$. The following result provides a complete characterization of weight functions.

**Theorem 2.2:** Let $\mathcal{H} = \{h_1, \ldots, h_m\}$ and $\mathcal{O} = \{ob_1, \ldots, ob_n\}$, and let $f$ be a real-valued function with domain $\mathcal{O} \times \mathcal{H}$ such that $f(ob, h) \in [0, 1]$. Then there exists an evidence space $\mathcal{E} = (\mathcal{H}, \mathcal{O}, \mu_{h_1}, \ldots, \mu_{h_m})$ such that $f = w_\mathcal{E}$ if and only if $f$ satisfies the following properties:

**WF1.** for every $ob \in \mathcal{O}$, $f(ob, \cdot)$ is a probability measure on $\mathcal{H}$,

**WF2.** there exists $x_1, \ldots, x_n \geq 0$ such that, for all $h \in \mathcal{H}$, $\sum_{i=1}^n f(ob_i, h)x_i = 1$.

This characterization is fundamental to the completeness of the axiomatization of the logic we introduce in the next section.

## 3 Reasoning about evidence

We introduce a logic $\mathcal{L}^{fo\text{-}ev}$ for reasoning about evidence, inspired by a logic introduced in FHM. The logic has both propositional features and first-order features. We take the probability of propositions, and view evidence as a proposition. On the other hand, we allow first-order quantification over numerical quantities, such as probabilities and evidence. The logic essentially considers two time periods, which can be thought of as the time before an observation is made and the time after an observation is made. For simplicity, we assume that exactly one observation is made. Thus, we can talk of the probability of a formula $\varphi$ before an observation is made, denoted $\Pr^0(\varphi)$, the probability of $\varphi$ after the observation, denoted $\Pr(\varphi)$, and the evidence of the observation $ob$ for a hypothesis $h$, denoted $we(ob, h)$. Of course, we want to be able to use the logic to relate all these quantities.

Formally, we start with two sets of primitive propositions, $\Phi_h = \{h_1, \ldots, h_{n_h}\}$ representing the hypotheses, and $\Phi_o = \{ob_1, \ldots, ob_{n_o}\}$ representing the observations. Let $\mathcal{L}_h(\Phi_h)$ be the propositional sublanguage of *hypothesis formulas* obtained by taking primitive propositions in $\Phi_h$ and closing off under negation and conjunction; we use $\rho$ to range over formulas of that sublanguage.

A *basic term* has the form $\Pr^0(\rho)$, $\Pr(\rho)$, or $we(ob, \rho)$, where $\rho$ is an hypothesis formula, and $ob$ is an observation. A *polynomial term* has the form $\theta_1 t_1 + \cdots + \theta_n t_n$, where $\theta_i$, $i = 1, \ldots, n$, is a real number and $t_i$ is a product of basic terms and variables (which intuitively range over the reals). A *polynomial inequality formula* has the form $p \geq \alpha$, where $p$ is a polynomial term and $\alpha$ is a real number. Let $\mathcal{L}^{fo\text{-}ev}(\Phi_h, \Phi_o)$ be the language obtained by starting out with the primitive propositions in $\Phi_h$ and $\Phi_o$ and polynomial inequality formulas, and closing off under conjunction, negation, and first-order quantification.

We use obvious abbreviations where needed, such as $\varphi \vee \psi$ for $\neg(\neg\varphi \wedge \neg\psi)$, $\varphi \Rightarrow \psi$ for $\neg\varphi \vee \psi$, $\exists x\varphi$ for $\neg\forall x(\neg\varphi)$, $\Pr(\varphi) - \Pr(\psi) \geq \alpha$ for $\Pr(\varphi) + (-1)\Pr(\psi) \geq \alpha$, $\Pr(\varphi) \geq \Pr(\psi)$ for $\Pr(\varphi) - \Pr(\psi) \geq 0$, $\Pr(\varphi) \leq \alpha$ for $-\Pr(\varphi) \geq -\alpha$, $\Pr(\varphi) < \alpha$ for $\neg(\Pr(\varphi) \geq \alpha)$, and $\Pr(\varphi) = \alpha$ for $(\Pr(\varphi) \geq \alpha) \wedge (\Pr(\varphi) \leq \alpha)$ (and analogous abbreviations for inequalities involving $\Pr^0$ and we).

**Example 3.1:** Consider again the situation given in Example 2.1. Let $\Phi_o$, the observations, consist of primitive propositions of the form heads$[m]$, where $m$ is an integer with $0 \leq m \leq 100$, indicating that $m$ heads out of 100 tosses have appeared. Let $\Phi_h$ consist of the two primitive propositions fair and doubleheaded. The computations in Example 2.1 can be written as follows:

$$\text{we}(\text{heads}[100], \text{fair}) = 1/(1+2^{100}) \wedge$$
$$\text{we}(\text{heads}[100], \text{doubleheaded}) = 2^{100}/(1+2^{100}).$$

We can also represent the rule of combination, for a fixed prior probability $\alpha$:

$$\text{Pr}^0(\text{fair}) = \alpha \wedge$$
$$\text{we}(\text{heads}[100], \text{fair}) = 1/(1+2^{100}) \Rightarrow$$
$$\text{Pr}(\text{fair}) = \alpha/(\alpha + (1-\alpha)2^{100}).$$

We will develop a deductive system to derive such conclusions in the next section. ∎

Now we consider the semantics. As usual, a model is a set of possible worlds. A world describes which hypothesis is true and which observation was made (recall that we have assumed that exactly one hypothesis is true, and exactly one observation is made), together with a probability distribution describing the prior probability, which is used to interpret $\text{Pr}^0$. Thus, a world has the form $(h, ob, \mu)$, where $h$ is a hypothesis, $ob$ is an observation, and $\mu$ is a probability distribution on $\Phi_h$. In addition, to interpret we, we need an evidence space over $\Phi_h$ and $\Phi_o$, which gives a probability measure $\mu_h$ on $\Phi_o$ for each hypothesis $h \in \Phi_h$. Thus, we take an *evidential structure* $M$ to be a tuple $(S \times \mathcal{P}, \mathcal{E})$, where $S \subseteq \Phi_h \times \Phi_o$, $\mathcal{P}$ is a set of probability distributions on $\Phi_h$, and $\mathcal{E}$ is an evidence space over $\Phi_h$ and $\Phi_o$. Note that the states of the structure are required to be only a subset of $\Phi_h \times \Phi_o$. Roughly speaking, this allows us to encode a priori information about the particular situation within the structure. Typically, such a priori information will rule out a particular combination of hypothesis and observation.

To interpret propositional formulas in $\mathcal{L}_h(\Phi_h)$, we define $[\![\rho]\!]$ to be the set of hypotheses denoted by the hypothesis formula $\rho$. Formally, $[\![true]\!] = \Phi_h$; $[\![h]\!] = \{h\}$; $[\![\neg\rho]\!] = \Phi_h - [\![\rho]\!]$; and $[\![\rho_1 \wedge \rho_2]\!] = [\![\rho_1]\!] \cap [\![\rho_2]\!]$.

In order to ascribe a semantics to first-order formulas that may contain variables, we need a valuation $v$ that assigns a real number to every variable. Given a valuation $v$, an evidential structure $M = (S \times \mathcal{P}, \mathcal{E})$, and a world $w = (h, ob, \mu)$, we can assign to a polynomial term $t$ a real number $[t]^{M,w,v}$ in a straightforward way:

$$[x]^{M,w,v} = v(x)$$
$$[\text{Pr}^0(\rho)]^{M,w,v} = \mu([\![\rho]\!])$$
$$[\text{Pr}(\rho)]^{M,w,v} = (\mu \oplus w_\mathcal{E}(ob, \cdot))([\![\rho]\!])$$
$$[\text{we}(ob', \rho)]^{M,w,v} = w_\mathcal{E}(ob', [\![\rho]\!])$$

$$[t_1 t_2]^{M,w,v} = [t_1]^{M,w,v} \times [t_2]^{M,w,v}$$
$$[\theta t]^{M,w,v} = \theta [t]^{M,w,v}$$
$$[t_1 + t_2]^{M,w,v} = [t_1]^{M,w,v} + [t_2]^{M,w,v}.$$

We define what it means for a formula $\varphi$ to be true (or satisfied) at a world $w$ of an evidential structure $M = (S \times \mathcal{P}, \mathcal{E})$ under valuation $v$, written $(M, w, v) \models \varphi$, as follows:

$(M, w, v) \models h$ if $w = (h, ob, \mu)$ for some $ob, \mu$

$(M, w, v) \models ob$ if $w = (h, ob, \mu)$ for some $h, \mu$

$(M, w, v) \models \neg \varphi$ if $(M, w, v) \not\models \varphi$

$(M, w, v) \models \varphi \wedge \psi$ if $(M, w, v) \models \varphi$ and $(M, w, v) \models \psi$

$(M, w, v) \models t \geq \alpha$ if $[t]^{M,w,v} \geq \alpha$

$(M, w, v) \models \forall x \varphi$ if $(M, w, v') \models \varphi$ for all valuations $v'$ that agree with $v$ on all variables but $x$.

If $(M, w, v) \models \varphi$ is true for all $v$, we write simply $(M, w) \models \varphi$. It is easy to check that if $\varphi$ is a closed formula (that is, one with no free variables), then $(M, w, v) \models \varphi$ if and only if $(M, w, v') \models \varphi$, for all $v, v'$. Therefore, given a closed formula $\varphi$, if $(M, w, v) \models \varphi$, then in fact $(M, w) \models \varphi$. We will typically be concerned only with closed formulas. If $(M, w) \models \varphi$ for all worlds $w$, then we write $M \models \varphi$ and say $\varphi$ is valid in $M$. Finally, if $M \models \varphi$ for all evidential structures $M$, we write $\models \varphi$ and say $\varphi$ is valid. In the next section, we will characterize axiomatically all the valid formulas of the logic.

## 4 Axiomatizing evidence

In this section we present a sound and complete axiomatization for our logic. To establish this, recall the following standard notions. We say a formula $\varphi$ is *provable* if it can be proven using the axioms and rules of inferences of that axiom system. An axiom system is *sound* if every provable formula is valid. An axiom system is *complete* if every valid formula is provable.

Our axiom system **AX** can be divided into four parts. The first set of axioms accounts for propositional reasoning:

**Taut.** All instances of valid formulas of first-order logic with equality.

**MP.** From $\varphi$ and $\varphi \Rightarrow \psi$ infer $\psi$.

Instances of **Taut** include, for example, all formulas of the form $\varphi \vee \neg \varphi$, where $\varphi$ is an arbitrary formula of the logic. Axiom **Taut** can be replaced by a sound and complete axiomatization for first-order logic with equality, as given, for instance, in Shoenfield [1967] or Enderton [1972].

The second set of axioms accounts for reasoning about polynomial inequalities:

**Ineq**. All instances of valid formulas about real closed fields, with nonlogical symbols $+, \cdot, <, 0, 1, -1$.

Valid formulas about real closed fields include, for example, the fact that addition on the reals is associative, $\forall x \forall y \forall z ((x+y)+z = x+(y+z))$, or 1 being the identity for multiplication, $\forall x (x \cdot 1 = x)$. As for **Taut**, we could replace **Ineq** by a sound and complete axiomatization for real closed fields (cf. [Fagin, Halpern, and Megiddo 1990; Shoenfield 1967; Tarski 1951]).

The third set of axioms essentially captures the fact that there is a single hypothesis and a single observation that holds per state.

**H1**. $h_1 \vee \ldots \vee h_{n_h}$.

**H2**. $h_i \Rightarrow \neg h_j$ if $i \neq j$.

**O1**. $ob_1 \vee \ldots \vee ob_{n_o}$.

**O2**. $ob_i \Rightarrow \neg ob_j$ if $i \neq j$.

The last set of axioms concerns reasoning about probabilities and evidence proper. The axioms for probability are taken from FHM.

**Pr1**. $\Pr^0(true) = 1$.

**Pr2**. $\Pr^0(\rho) \geq 0$.

**Pr3**. $\Pr^0(\rho_1 \wedge \rho_2) + \Pr^0(\rho_1 \wedge \neg \rho_2) = \Pr^0(\rho_1)$.

**Pr4**. $\Pr^0(\rho_1) = \Pr^0(\rho_2)$ if $\rho_1 \Leftrightarrow \rho_2$ is a propositional tautology.

Axiom **Pr1** simply say that the event *true* has probability 1. Axiom **Pr2** says that probability is nonnegative. Axiom **Pr3** captures finite additivity. It is not possible to express countable additivity in our logic. On the other hand, we do not need countable additivity. Roughly speaking, as we establish in the next section, if a formula is satisfiable at all, it is satisfiable in a finite structure. Similar axioms capture posterior probability formulas:

**Po1**. $\Pr(true) = 1$.

**Po2**. $\Pr(\rho) \geq 0$.

**Po3**. $\Pr(\rho_1 \wedge \rho_2) + \Pr(\rho_1 \wedge \neg \rho_2) = \Pr(\rho_1)$.

**Po4**. $\Pr(\rho_1) = \Pr(\rho_2)$ if $\rho_1 \Leftrightarrow \rho_2$ is a propositional tautology.

Finally, we need axioms to account for the behavior of the evidence operator we. What are these properties? For one thing, the weight function acts like a probability on hypotheses, for each fixed observation. This gives the following four axioms:

**E1**. $\text{we}(ob, true) = 1$.

**E2**. $\text{we}(ob, \rho) \geq 0$.

**E3**. $\text{we}(ob, \rho_1 \wedge \rho_2) + \text{we}(ob, \rho_1 \wedge \neg \rho_2) = \text{we}(ob, \rho_1)$.

**E4**. $\text{we}(ob, \rho_1) = \text{we}(ob, \rho_2)$ if $\rho_1 \Leftrightarrow \rho_2$ is a propositional tautology.

Second, evidence connects the prior and posterior beliefs via Dempster's Rule of Combination, as in (2). This is captured by the following axiom. (Note that, since we do not have division in the language, we crossmultiply to clear the denominator.)

**E5**. $ob \Rightarrow (\Pr^0(h)\text{we}(ob, h) =$
$\Pr(h)\Pr^0(h_1)\text{we}(ob, h_1) + \ldots +$
$\Pr(h)\Pr^0(h_{n_h})\text{we}(ob, h_{n_h}))$.

This is not quite enough. As we saw in Section 2, it is not sufficient for a function on observations and hypotheses to act as a probability measure on the hypotheses for each observation in order to be a weight of evidence function. Property **WF2** in Theorem 2.2 is required for a function to be an evidence function. The following axiom captures **WF2** in our logic:

**E6**. $\exists x_1 \ldots \exists x_{n_o} (x_1 \geq 0 \wedge \ldots \wedge x_{n_o} \geq 0$
$\wedge \text{we}(ob_1, h_1)x_1 + \ldots +$
$\text{we}(ob_{n_o}, h_1)x_{n_o} = 1$
$\wedge \ldots$
$\wedge \text{we}(ob_1, h_{n_h})x_1 + \ldots +$
$\text{we}(ob_{n_o}, h_{n_h})x_{n_o} = 1)$.

Note that axiom **E6** is the only axiom that requires quantification.

**Theorem 4.1**: *AX is a sound and complete axiomatization for $\mathcal{L}^{fo-ev}(\Phi_h, \Phi_o)$ with respect to evidential structures.*

As usual, soundness is straightforward, and to prove completeness, it suffices to show that if a formula $\varphi$ is consistent with **AX**, it is satisfiable in an evidential structure. However, the usual approach for proving completeness in modal logic, which involves considering maximal consistent sets and canonical structures does not work. The problem is that there are maximal consistent sets of formulas that are not satisfiable. For example, there is a maximal consistent set of formulas that includes $\Pr(\rho) > 0$ and $\Pr(\rho) \leq 1/n$ for $n = 1, 2, \ldots$. This is clearly unsatisfiable. Our proof follows the techniques developed in FHM.

To express axiom **E6**, we needed to have quantification in the logic. An interesting question is whether it is possible to give a sound and complete axiomatization to the propositional fragment of our logic (without quantification or variables). To do this, we need to give quantifier-free axioms to replace axiom **E6**. This amounts to asking whether there is a simpler property than **WF2** in Theorem 2.2 that characterizes weight of evidence functions. This remains an open question.

## 5  Decision procedures

In this section, we consider the decision problem for our logic, that is, the problem of deciding whether a given formula $\varphi$ is satisfiable. In order to precisely state the problem, however, we need to highlight a particular subtlety of the logic. The language of the logic is parametrized by sets of primitive propositions $\Phi_h$ and $\Phi_o$. In other words, we actually have a family of logics, indexed by $\Phi_h$ and $\Phi_o$. In most logics, there is a certain monotonicity property that says that the choice of underlying primitive propositions is essentially irrelevant. For example, if a propositional formula $\varphi$ that contains only primitive propositions in some set $\Phi$ is true with respect to all truth assignments to $\Phi$, then it remains true with respect to all truth assignments to any set $\Phi' \supseteq \Phi$. This monotonicity property does not hold here. For example, axiom **H1** clearly depends on the set of hypotheses and observations; it is no longer valid if the set is changed. The same is true for **O1**. Axioms **E5** and **E6** also depend on the set of hypotheses and observations. To deal with this, we assume that the satisfiability algorithm gets as input $\Phi_h$, $\Phi_o$, and a formula $\varphi \in \mathcal{L}^{fo\text{-}ev}(\Phi_h, \Phi_o)$.

Because $\mathcal{L}^{fo\text{-}ev}(\Phi_h, \Phi_o)$ contains the full theory of real closed field, it is unsurprisingly difficult to decide. For our decision procedure, we can use any of the recent doubly-exponential procedures that have been developed for the decision problem of the theory of real closed field [Renegar 1992; Basu 1999].

**Theorem 5.1:** *There is a procedure that runs in doubly-exponential time for deciding whether a formula $\varphi$ of $\mathcal{L}^{fo\text{-}ev}(\Phi_h, \Phi_o)$ is satisfiable in an evidential structure.*

This is essentially the best we can do; Weispfenning [1988] shows a doubly-exponential lower bound for the decision problem in the theory of real closed fields.

The main culprit for the doubly exponential-time complexity is the theory of real closed fields, which we had to add to the logic to be able to even write down axiom **E6** of the axiomatization **AX**. (Recall that axiom **E6** requires quantification.) However, if we are not interested in axiomatizations, but simply in verifying properties of probabilities and weights of evidence, we can consider the following propositional (quantifier-free) fragment of our logic, which we call $\mathcal{L}^{ev}(\Phi_h, \Phi_o)$. This sublanguage essentially allows linear inequality formulas, without the use of variables or quantifiers. For instance, a linear inequality formula takes the form $\Pr^0(\rho) + 3\text{we}(ob, \rho) + 5\Pr(\rho') \geq 7$. This is sufficiently expressive to express many properties of interest; for instance, it can certainly express the relationship between prior probability and posterior probability through the weight of evidence of a particular observation, as shown in Example 3.1. Reasoning about the propositional fragment of our logic $\mathcal{L}^{ev}(\Phi_h, \Phi_o)$ is easier than the full language:

**Theorem 5.2:** *The problem of deciding whether a formula $\varphi$ of $\mathcal{L}^{ev}(\Phi_h, \Phi_o)$ is satisfiable in an evidential structure is NP-complete.*

Since $\mathcal{L}^{ev}(\Phi_h, \Phi_o)$ allows only linear inequalities, it cannot express the general connection between priors, posteriors, and evidence captured by axiom **E5**. As observed in FHM, we can extend $\mathcal{L}^{ev}$ to allow multiplication of probability terms. This leads to a more expressive logic that can be decided in polynomial space, using Canny's [1988] procedure for deciding the validity of quantifier-free formulas in the theory of real closed fields.

## 6  Conclusion

We have presented in this paper a logic of evidence that can capture in a logical form the relationship between the prior probability of hypotheses, the weight of evidence of particular observations, and the posterior probability of hypotheses after the observations. The key aspect of our logic is that it allows reasoning about the weights of evidence independently of the prior probabilities.

The evidential structures we have considered in this paper are essentially static, in that they model only the situation where a single observation is made. This let us focus on the relationship between the prior and posterior probabilities on hypotheses and the weight of evidence of a single observation. In a related paper [Halpern and Pucella 2003], we consider evidence in the context of randomized algorithms; we use evidence to characterize the information provided by, for example, a randomized algorithm for primality when it says that a number is prime. The framework in [Halpern and Pucella 2003] is dynamic; sequences of observations are made over time.

It is straightforward to extend the framework presented here to a dynamic setting. Rather than just considering states, we consider sequences of states (or runs), representing the evolution of the system over time. The models are now sets of runs, with a set of prior probabilities on the hypotheses that hold in the runs. We can then modify the logic to express properties of evidence in this more dynamic setting. In some ways considering a dynamic setting simplifies things. Rather than talking about the prior and posterior probability using different operators, we need only a single probability operator that represents the probability of an hypothesis at the current time. To express the analogue of axiom **E5** in this logic, we need to be able to talk about the probability at the next time step. This can be done by adding the "next-time" operator $\bigcirc$ to the logic, where $\bigcirc \varphi$ holds at the current time if $\varphi$ holds at the next time step. We can further extend the logic to talk about the weight of evidence of a sequence of observations. We leave details to the full paper.

**Acknowledgments**

We thank Dexter Kozen for useful discussions. This work was supported in part by NSF under grant CTC-0208535, by ONR under grants N00014-00-1-03-41 and N00014-01-10-511, by the DoD Multidisciplinary University Research Initiative (MURI) program administered by the ONR under grant N00014-01-1-0795, and by AFOSR under grant F49620-02-1-0101.

**References**

Basu, S. (1999). New results on quantifier elimination over real closed fields and applications to constraint databases. *Journal of the ACM 46*(4), 537–555.

Canny, J. F. (1988). Some algebraic and geometric computations in PSPACE. In *Proc. 20th ACM Symp. on Theory of Computing*, pp. 460–467.

Carnap, R. (1962). *Logical Foundations of Probability* (Second ed.). University of Chicago Press.

de Alfaro, L. (1998). *Formal Verification of Probabilistic Systems*. Ph. D. thesis, Stanford University. Available as Technical Report STAN-CS-TR-98-1601.

Enderton, H. B. (1972). *A Mathematical Introduction to Logic*. Academic Press.

Fagin, R. and J. Y. Halpern (1994). Reasoning about knowledge and probability. *Journal of the ACM 41*(2), 340–367.

Fagin, R., J. Y. Halpern, and N. Megiddo (1990). A logic for reasoning about probabilities. *Information and Computation 87*(1/2), 78–128.

Fischer, M. J. and L. D. Zuck (1988). Reasoning about uncertainty in fault-tolerant distributed systems. Technical Report YALEU/DCS/TR–643, Yale University.

Good, I. J. (1950). *Probability and the Weighing of Evidence*. Charles Griffin & Co. Ltd.

Good, I. J. (1960). Weights of evidence, corroboration, explanatory power, information and the utility of experiments. *Journal of the Royal Statistical Society, Series B 22*, 319–331.

Halpern, J. Y. and R. Fagin (1992). Two views of belief: belief as generalized probability and belief as evidence. *Artificial Intelligence 54*, 275–317.

Halpern, J. Y., Y. Moses, and M. R. Tuttle (1988). A knowledge-based analysis of zero knowledge. In *Proc. 20th ACM Symp. on Theory of Computing*, pp. 132–147.

Halpern, J. Y. and R. Pucella (2003). Probabilistic algorithmic knowledge. In *Theoretical Aspects of Rationality and Knowledge: Proc. Ninth Conference (TARK 2003)*, pp. 118–130.

Halpern, J. Y. and M. R. Tuttle (1993). Knowledge, probability, and adversaries. *Journal of the ACM 40*(4), 917–962.

Kyburg, H. (1983). Recent work in inductive logic. In T. Machan and K. Lucey (Eds.), *Recent Work in Philosophy*, pp. 87–150. Rowman & Allanheld.

Larsen, K. and A. Skou (1991). Bisimulation through probabilistic testing. *Information and Computation 94*(1), 1–28.

Milne, P. (1996). $\log[p(h|eb)/p(h|b)]$ is the one true measure of confirmation. *Philosophy of Science 63*, 21–26.

Popper, K. R. (1959). *The Logic of Scientific Discovery*. Hutchinson.

Renegar, J. (1992). On the computational complexity and geometry of the first order theory of the reals. *Journal of Symbolic Computation 13*(3), 255–352.

Shafer, G. (1976). *A Mathematical Theory of Evidence*. Princeton University Press.

Shafer, G. (1982). Belief functions and parametric models (with commentary). *Journal of the Royal Statistical Society, Series B 44*, 322–352.

Shoenfield, J. R. (1967). *Mathematical Logic*. Addison-Wesley.

Tarski, A. (1951). *A Decision Method for Elementary Algebra and Geometry* (2nd ed.). Univ. of California Press.

van Glabbeek, R., S. Smolka, B. Steffen, and C. Tofts (1990). Reactive, generative, and stratified models of probabilistic processes. In *Fifth Annual IEEE Symposium on Logic in Computer Science*, pp. 130–141. IEEE Computer Society Press.

Vardi, M. Y. (1985). Automatic verification of probabilistic concurrent finite-state programs. In *Proc. 26th IEEE Symp. on Foundations of Computer Science*, pp. 327–338.

Walley, P. (1987). Belief function representations of statistical evidence. *Annals of Statistics 18*(4), 1439–1465.

Weispfenning, V. (1988). The complexity of linear problems in fields. *Journal of Symbolic Computation 5*(1/2), 3–27.